# Improving Uncertainty Sampling with Bell Curve Weight Function

Zan-Kai Chong[1*], Hiroyuki Ohsaki[2], Bok-Min Goi[3]

[1] Independent Researcher, Malaysia.
[2] School of Science and Technology, Kwansei Gakuin University, Japan.
[3] Lee Kong Chian Faculty of Engineering Science, Universiti Tunku Abdul Rahman, Malaysia.

* Corresponding author. Email: zankai@ieee.org

**Abstract:** Typically, a supervised learning model is trained using passive learning by randomly selecting unlabelled instances to annotate. This approach is effective for learning a model, but can be costly in cases where acquiring labelled instances is expensive. For example, it can be time-consuming to manually identify spam mails (labelled instances) from thousands of emails (unlabelled instances) flooding an inbox during initial data collection. Generally, we answer the above scenario with uncertainty sampling, an active learning method that improves the efficiency of supervised learning by using fewer labelled instances than passive learning. Given an unlabelled data pool, uncertainty sampling queries the labels of instances where the predicted probabilities, $p$, fall into the uncertainty region, i.e., $p \approx 0.5$. The newly acquired labels are then added to the existing labelled data pool to learn a new model. Nonetheless, the performance of uncertainty sampling is susceptible to the Area of Unpredictable Responses (AUR) and the nature of the dataset. It is difficult to determine whether to use passive learning or uncertainty sampling without prior knowledge of a new dataset. To address this issue, we propose bell curve sampling, which employs a bell curve weight function to acquire new labels. With the bell curve centred at $p = 0.5$, bell curve sampling selects instances whose predicted values are in the uncertainty area most of the time without neglecting the rest. Simulation results show that, most of the time bell curve sampling outperforms uncertainty sampling and passive learning in datasets of different natures and with AUR.

## 1. Introduction

Supervised learning is a sub-field of machine learning, and it trains a model to learn the relationship between input and output sets through labelled instances [1]. With the quick expanding of digital era, supervised learning has been widely applied in many practical applications that amass large datasets such as spam detection, face detection, voice recognition, etc. [2].

In some cases, it may take little effort to acquire a large number of unlabelled instances, but labelling them turns out to be laborious and costly. For example, credit scoring models for microfinance businesses are normally trained with loan repayment behaviour datasets and it may take a few months to conclude a borrower as a bad payer. In particular, a finance institute may receive thousands of loan applications (unlabelled instances) regularly, but it takes substantial risk to approve loans [3, 4] and a lengthy observation to identify bad payers (labelled instances).

Cohn and Atlas *et al.* [5] have long been aware of the potential of active learning in addressing expensive acquisition cost of labelled samples in passive learning. The process to select instances to label is named as "query" in active learning and it enables a model to learn with a lesser number of labelled instances efficiently as compared to passive learning. We consider a popular pool-based active learning, namely uncertainty sampling [6, 7] in this paper. Given a data set of balanced dichotomous responses, uncertainty sampling queries the labels of instances in the unlabelled data pool, in which their predicted probabilities fall into the uncertainty region ( $p \approx 0.5$) [8].

Uncertainty sampling has gained decent attention in the research community [6, 9] with some reported negative results [10–12]. They explain the results are due to the sampling bias causing the disparity in between the known feature distribution and the actual one [13]. Additionally, the inaccurate initial model also drives subsequent the improper instances selection that further drifting the intended sampling strategy [10].

In this paper, we assume the presence of an Area of Unpredictable Responses (AUR), i.e., a subset of dataset, where all predictions here yield random responses grievously due to unavoidable noises or weak predictive features as shown in the overlapped area of Fig. 1. The simulation results in Section 4 show that the performance of uncertainty sampling is precarious in datasets of different nature and AUR. It is indeterminate to determine the best sampling method without the prior knowledge about a new dataset.

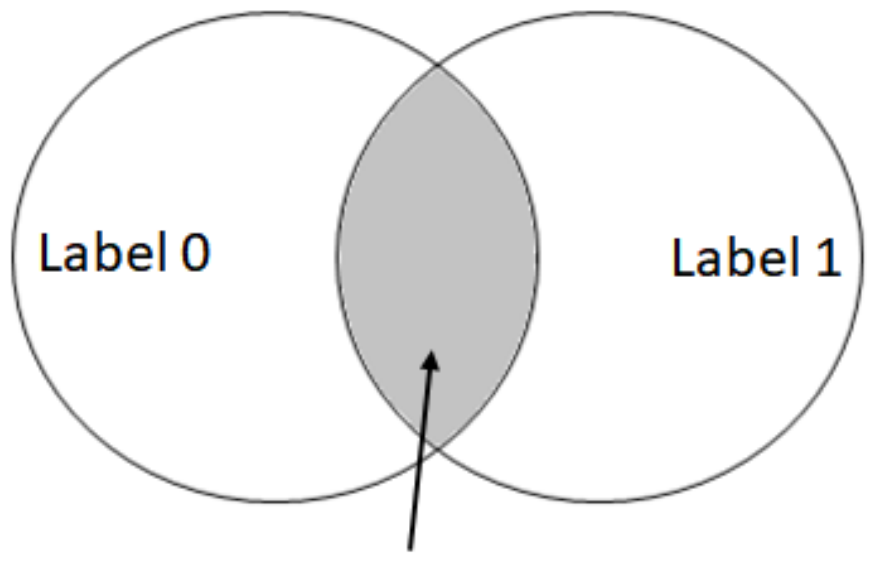

Fig. 1. The circles on the left and right contain instances of labels "0" and "1", respectively. The overlapped area in the middle is filled with instances of indeterminate responses.

We adduce the idea of heuristic optimization technique [14] that both intensification and diversification are necessary to search for a solution. Accordingly, we propose a sampling that applies a bell curve weight acquisition function, where instances are selected from the uncertainty region most of the time without neglecting the instances from other regions. Comparing with uncertainty sampling and passive learning, simulation results reported that bell curve learning aces the performance in the datasets of diversified AUR in general.

This paper is organised as following: We discuss the algorithm of uncertainty sampling with bare mathematical notations in Section 2. Section 3 presents the proposed bell curve sampling and its assumptions. Next, the simulation setup and performance of passive learning, uncertainty sampling and bell curve sampling are presented in Section 4. With that, we draw the conclusion in Section 5.

## 2. Uncertainty Sampling

Assume that we have a large finite set of tuples $D^{(\text{all})} = \left\{(x,\ y): x \in \Re^d,\ y \in \{0,1\}\right\}$ that describes the population of our interest with $x$ and $y$ represent $d$-dimensions feature vectors and dichotomous responses respectively. A relatively small and large numbers of tuples are taken from $D^{(\text{all})}$ and they are denoted as *known* data pool and *unknown* data pool, i.e., $D^{(\text{known})} = \{(x, y)\}$ and $D^{(\text{unknown})} = \{(x, \emptyset)\}$, where $\emptyset$ symbolises the masked responses, and $\left|D^{(\text{unknown})}\right| > \left|D^{(\text{known})}\right|$. In practical, we do not have test data to learn the model performance. However, in the simulation a test data pool, $D^{(\text{test})} = \{(x, y)\}$ from $D^{(\text{all})}$ will be given to evaluate the impact of the learning algorithms. For convenience, we use subscripts, e.g., $D_q^{(\cdot)}$ to denote the corresponding data pool at $q$-th query.

Referring an efficient machine learning algorithm as MLA, at the beginning of the query, we have a prior model that is trained from $D_1^{(\text{known})}$, i.e., $D_1^{(\text{known})} \xrightarrow{\text{MLA}} \text{model}_1$. The rest of steps go as follows.

- Step 1. With $\text{model}_1$, we make inference to all the instances in current $D_1^{(\text{unknown})}$ to yield the predicted outputs $\widetilde{y_1}$ , and the corresponding predicted probabilities $\widetilde{p_1}$ , i.e., $D_1^{(\text{unknown})} \xrightarrow{\text{model}_1} \widetilde{D_1}^{(\text{unknown})}$ and $\widetilde{D_1}^{(\text{unknown})} = \{(x_i, \emptyset, \widetilde{y_1}, \widetilde{p_1})\}$.
- Step 2. Uncertainty sampling selects $n$ instances from $\widetilde{D_1}^{(\text{unknown})}$, in which the values of corresponding $\tilde{p}$ in the uncertainty region. With $\emptyset$ be revealed during the annotation, $D_1^{(\text{query})} = \{(x, y): \emptyset \to y\}, D \subset \widetilde{D}_1^{(\text{unknown})}, \left|D_1^{(\text{query})}\right| = n$ and the selected instances will be removed from $D_1^{(\text{unknown})}$ i.e., $D_1^{(\text{unknown})} \setminus D_1^{(\text{query})} \to D_1^{(\text{unknown})}$ .
- Step 3. We merge the new labelled instances into the known data pool, i.e., $D_1^{(\text{query})} \cup D_1^{(\text{known})} \to D_1^{(\text{known})}$ .
- Step 4. Renaming the indices $D_1^{(\text{unknown})}$ to $D_2^{(\text{unknown})}$ and $D_1^{(\text{known})}$ to $D_2^{(\text{known})}$, we train a new model with $D_2^{(\text{known})} \xrightarrow{\text{MLA}} \text{model}_2$ . Then, the next query will be repeated from Step 1 with all the indices to be renamed correctly. The process is ended when the stopping criteria are met.

## 3. Bell Curve Sampling

This section elaborates the assumptions and the principle of bell curve sampling.

### 3.1. Assumptions

We assume datasets with balanced dichotomous responses (i.e., about the same amount of 0's and 1's in responses) with the presence of AUR. An observer has access to the feature vectors of each instance in an unknown data pool, but it has no knowledge about the true responses.

### 3.2. Weight-Probabilities Distribution

Fig. 2 (a) and (b) illustrate the weight-probabilities distributions for passive learning and uncertainty sampling to select the instances to annotate. The former selects instances randomly, i.e., all instances have equal weight to be selected regardless of their predicted probabilities. Meanwhile, the latter only selects the instances near to $p \approx 0.5$. As mentioned in Section 1, it is precarious to determine the best sampling method without the prior knowledge about the data set. With that, we propose *bell curve sampling* and it employs a bell curve weight distribution to marry the strength of both passive learning and uncertain sampling in coping datasets of diversified nature and AUR. As the name suggests, bell curve sampling contains a bell-curve-like weight distribution with peak at $p = 0.5$ as shown in Fig. 2 (c). Instances near to $p = 0.5$ will be chosen frequently without neglecting the instances from other regions. Such selection strategy follows the idea of intensification and diversification approaches in heuristic search.

To derive the bell curve weight-probabilities distribution, let $\text{Beta}(p, \alpha, \beta)$ be the probability density function of beta distribution for $0 \le p \le 1$ , the shape parameters $\alpha, \beta > 0$ and $\alpha, \beta, p \in \Re$. As demonstrated in Fig. 3, having $\alpha = \beta$ moves the centre of bell curve to $p = 0.5$, and higher values of $\alpha$ and $\beta$ transform the bell curve to be steeper with shorter width. For example, with $\alpha = \beta = 5$, 95% of area are covered in the probabilities range of (0.3920, 0.6080) as shown in Table 1. Meanwhile, the probabilities range shrinks to (0.4241, 0.5759) for $\alpha = \beta = 10$. In general, the performance of bell curve sampling is identical to passive learning with $\alpha = \beta = 1$ as the weights are distributed uniformly. Moreover, it acts as the uncertainty sampling with a high value of $\alpha$ and $\beta$ due to shorter probabilities range. We look for the reasonable values of $\alpha$ and $\beta$ that synergize both passive learning and uncertainty sampling in Section 4.

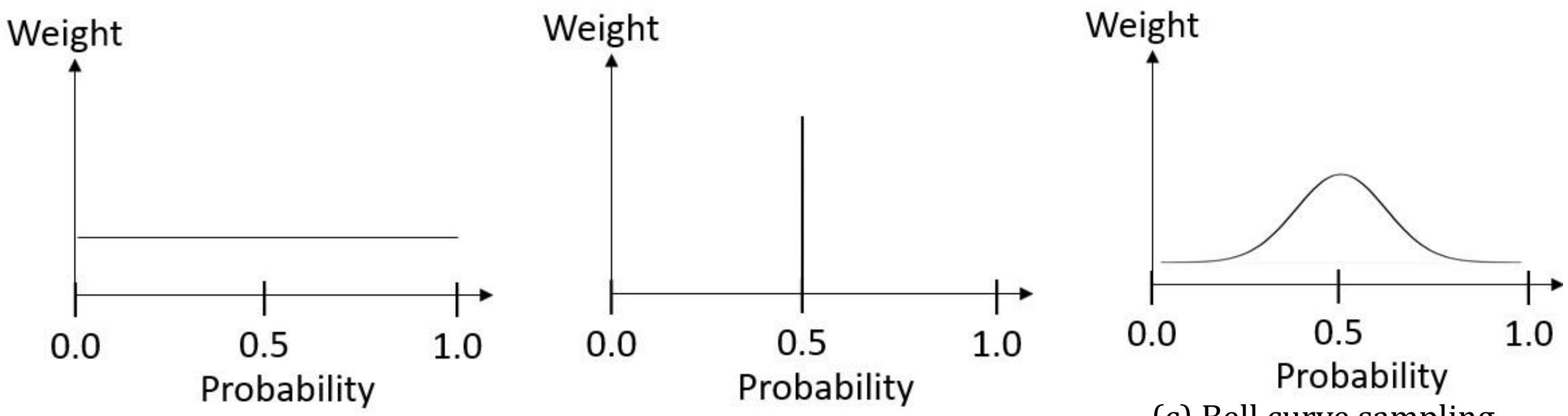

(a) Passive learning (b) Uncertainty sampling (c) Bell curve sampling

Fig. 2. Weight-probabilities distributions of various sampling methods.

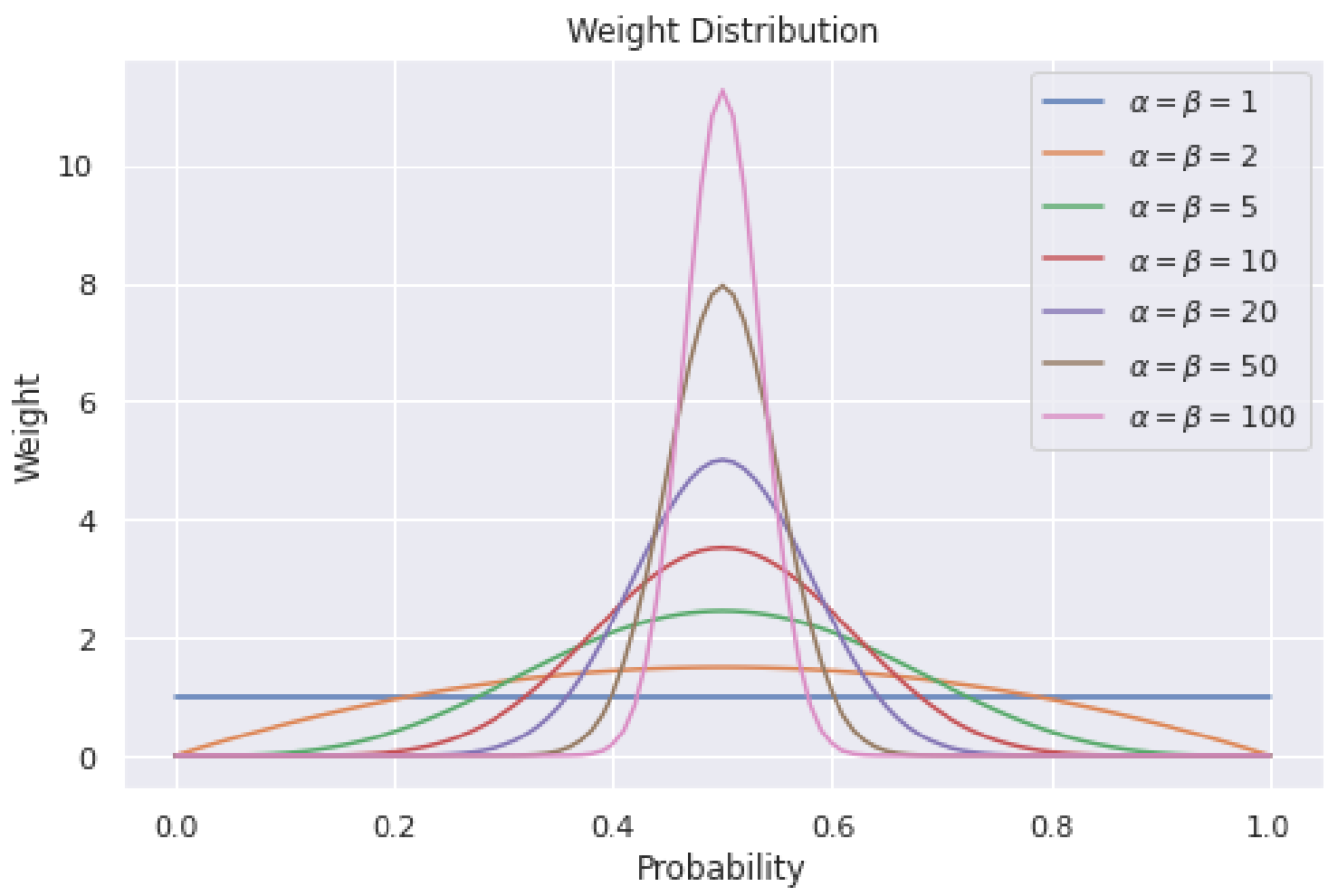

Fig. 3. Beta distribution of various pair of $\alpha = \beta$.

Table 1. Lower and Upper Bound of 95% Area under Curve for Various Set of $\alpha$ and $\beta$ in Beta Distribution

| $\alpha$ | $\beta$ | Lower | Upper |
|---|---|---|---|
| 2 | 2 | 0.3264 | 0.6736 |
| 5 | 5 | 0.3920 | 0.6080 |
| 10 | 10 | 0.4241 | 0.5759 |
| 20 | 20 | 0.4465 | 0.5535 |
| 50 | 50 | 0.4662 | 0.5338 |
| 100 | 100 | 0.4761 | 0.5239 |

## 4. Simulation

We study the performance of bell curve sampling with simulation. This section illustrates the simulation setup, the artificial datasets and the simulation results.

### 4.1. Simulation Setup

We generate four types of artificial datasets, i.e., classification, blobs, circles and moons in various degrees of AUR using Scikit-learn [15]. Examples of the corresponding pairwise bivariate distributions are shown in Fig. 4. Classification and blobs datasets consist of four feature columns, whereas circles and moons datasets have two feature columns only. All datasets employ balanced dichotomous responses.

Generally, a sufficient large population of data set will first be generated and further segregated into three chunks randomly, i.e., known data pool, $D^{(\text{known})}$ of 10 instances, an unknown data pool, $D^{(\text{unknown})}$ of 1000 instances and a test data pool, $D^{(\text{test})}$ of 1000 instances. Note that $D^{(\text{test})}$ will be used to assess the model performance in the last step of each query.

We follow the query process that is described in Section 2. We change the sampling function in Step 2 to passive learning (random sampling) and bell curve sampling according to the simulation objective. Each query will select $n = 5$ instances from $D^{(\text{unknown})}$ to annotate (unmask the true responses) and a simulation will make a total of 20 queries, and $20 \times n = 100$ or 10% of the instances from $D^{(\text{unknown})}$ will be added to $D^{(\text{known})}$ eventually. All the models are built with AutoML [16] using optimal parameters. No further feature engineering is conducted on the datasets.

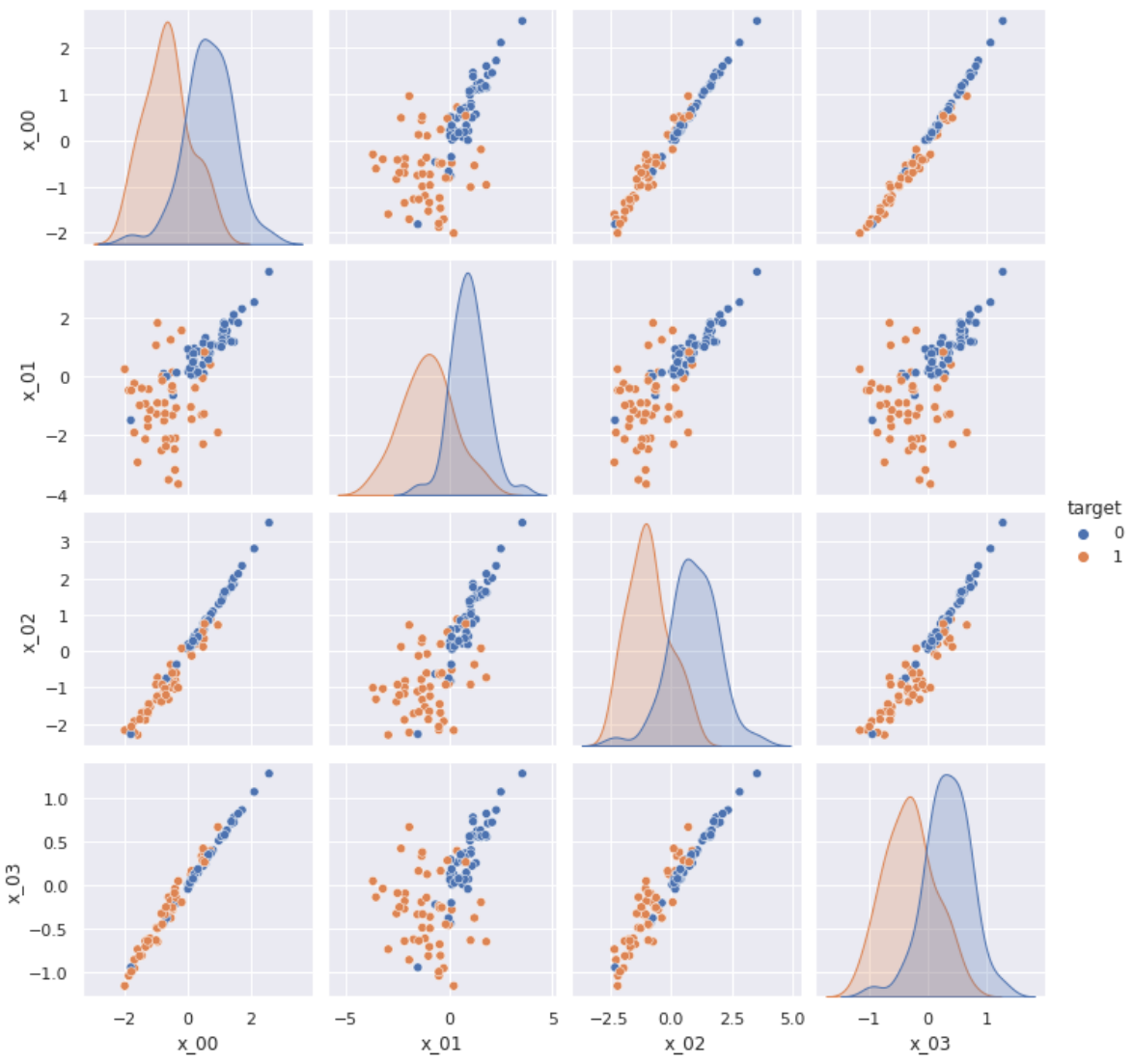

(a) Classification dataset

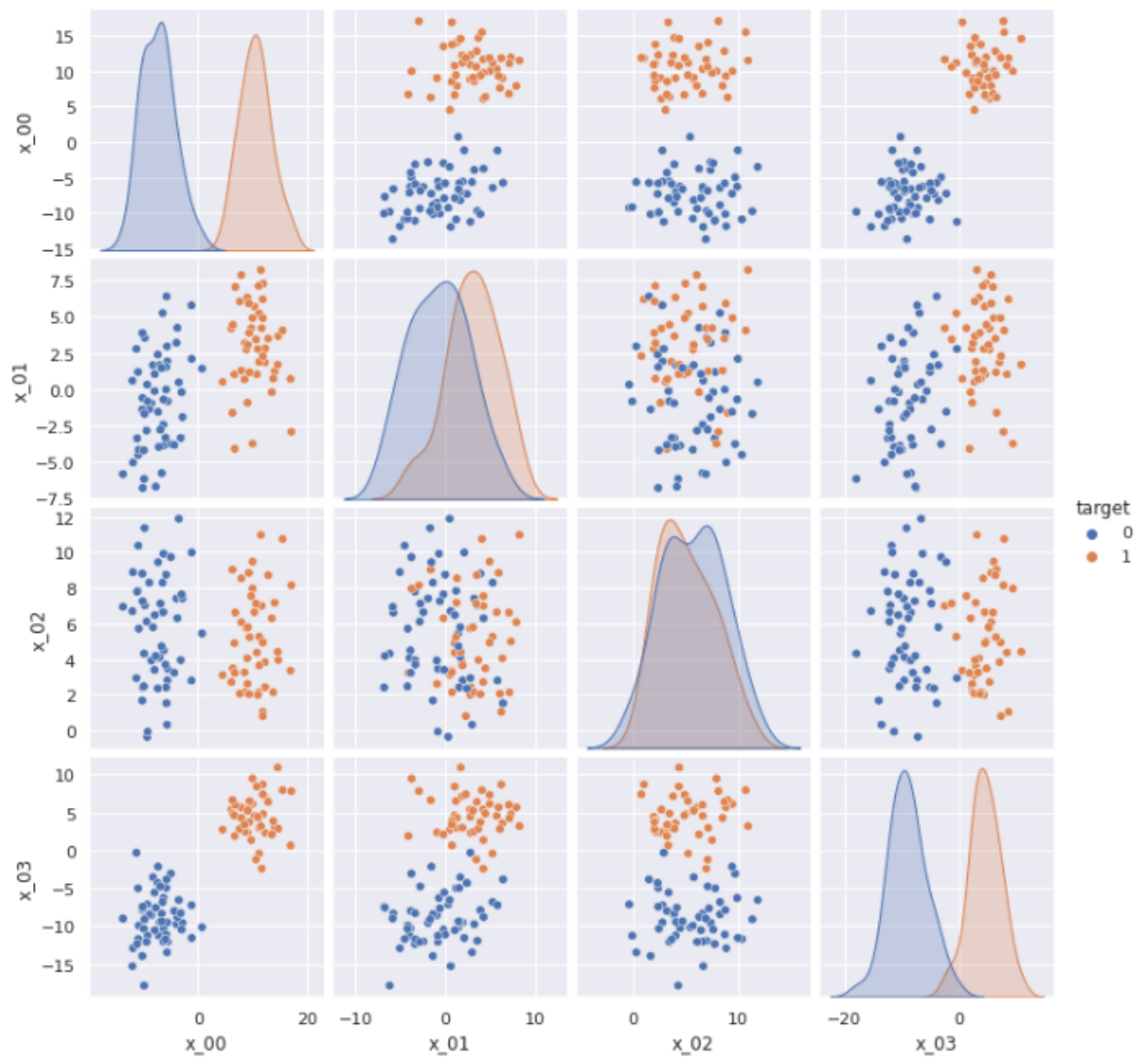

(b) Blobs dataset

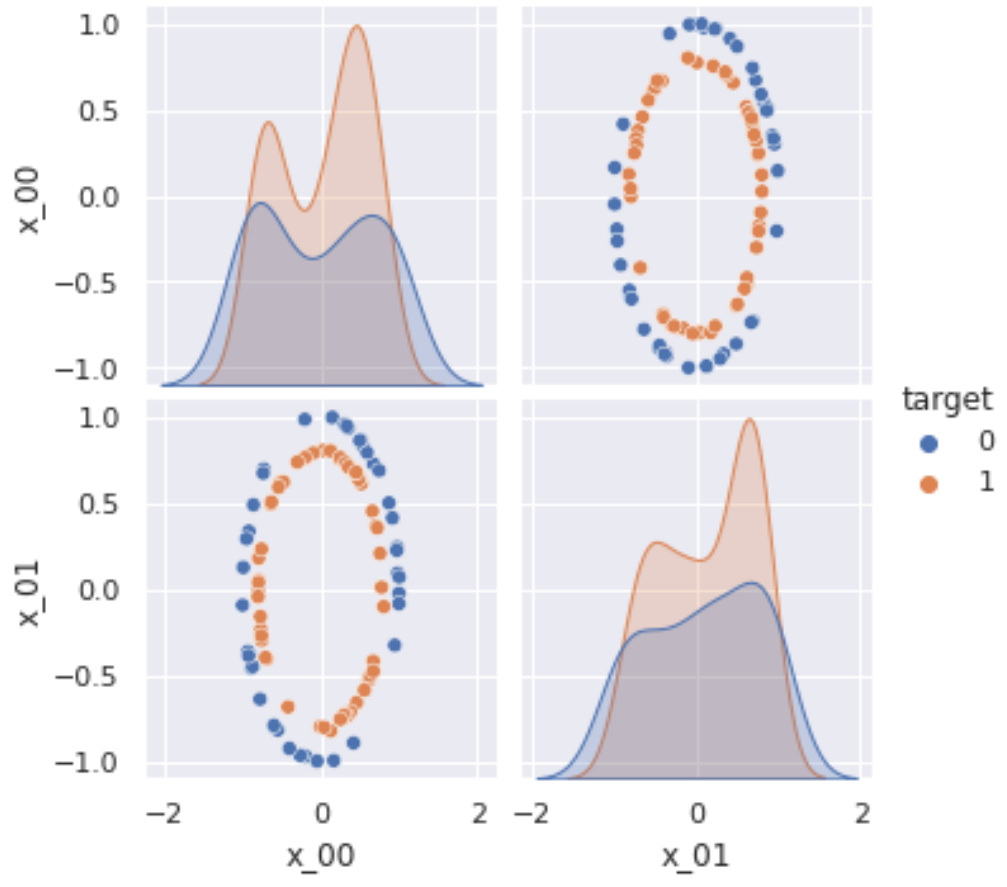

(c) Circles dataset

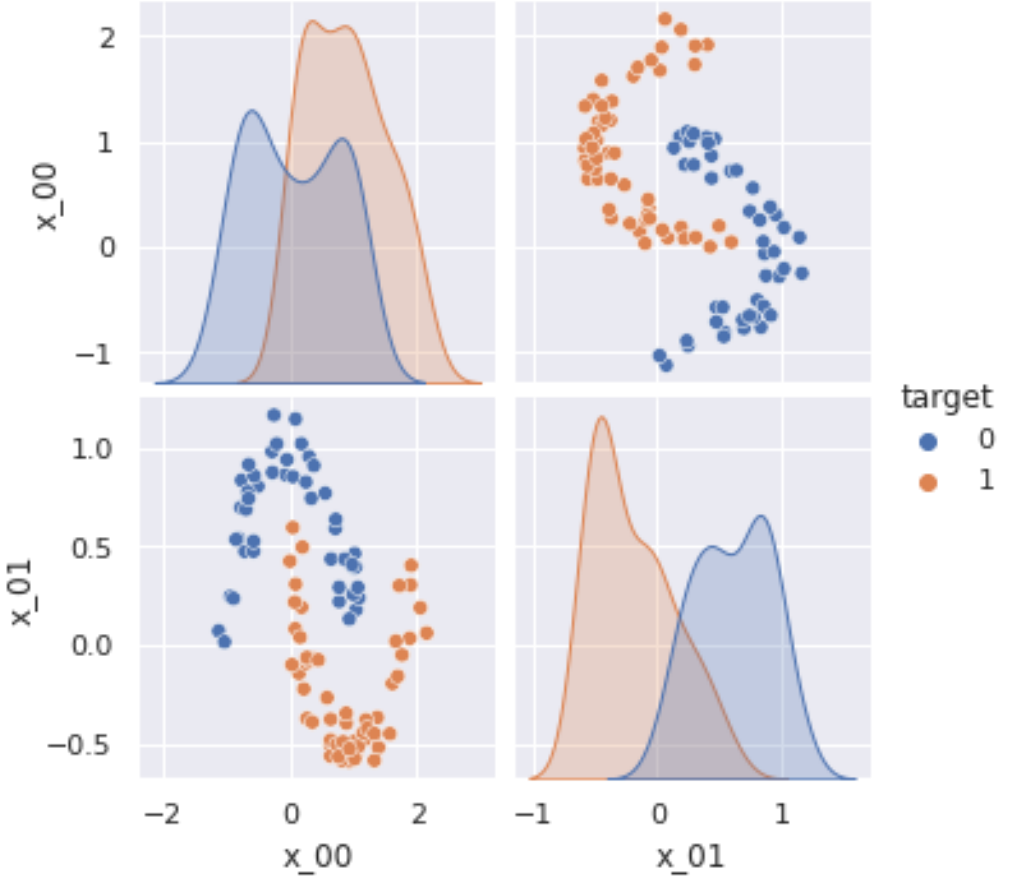

(d) Moons dataset

Fig. 4. Pairwise bivariate distributions of Scikit-learn's artificial datasets.

## 4.2. Results

Figs. 5–8 depict the performance of passive learning, uncertainty sampling and bell curve sampling in the classification, blobs, circles and moons datasets using $\alpha = \beta = 10$. We generated datasets with varying levels of AUR, namely (a) low, (b) median, and (c) high, by adjusting their respective parameters. For example, increasing the class separation factor facilitates the separation of responses in classification datasets.

As discussed in Section 1, our observations indicate that the performance of uncertainty sampling is dependent on the nature of the datasets and their AUR. For instance, uncertainty sampling demonstrated superior performance over passive learning in the blobs datasets with median and high AUR (Fig. 6 (b) and (c)), while it exhibited similar performance to passive learning in the low AUR case (Fig. 6 (a)). Conversely, passive learning outperformed uncertainty sampling in the circles and moons datasets (Figs. 7 and 8). Nonetheless, bell curve sampling generally exhibited better performance across most datasets. This approach showcased competitive performance as demonstrated in the best learner cases, such as Figs. 7(c) and 8(c), where passive learning was the top learner, followed by bell curve sampling and uncertainty sampling.

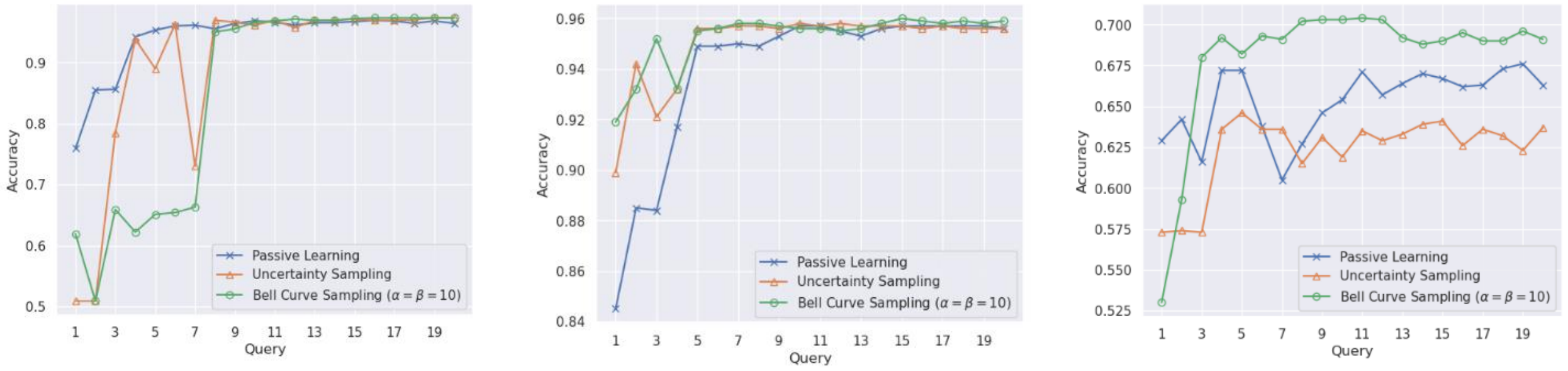

(a) Low AUR (b) Median AUR (c) High AUR

Fig. 5. Performance of various learning methods in classification datasets of class separation factors (a) 2.0, (b) 0.8 and (c) 0.3.

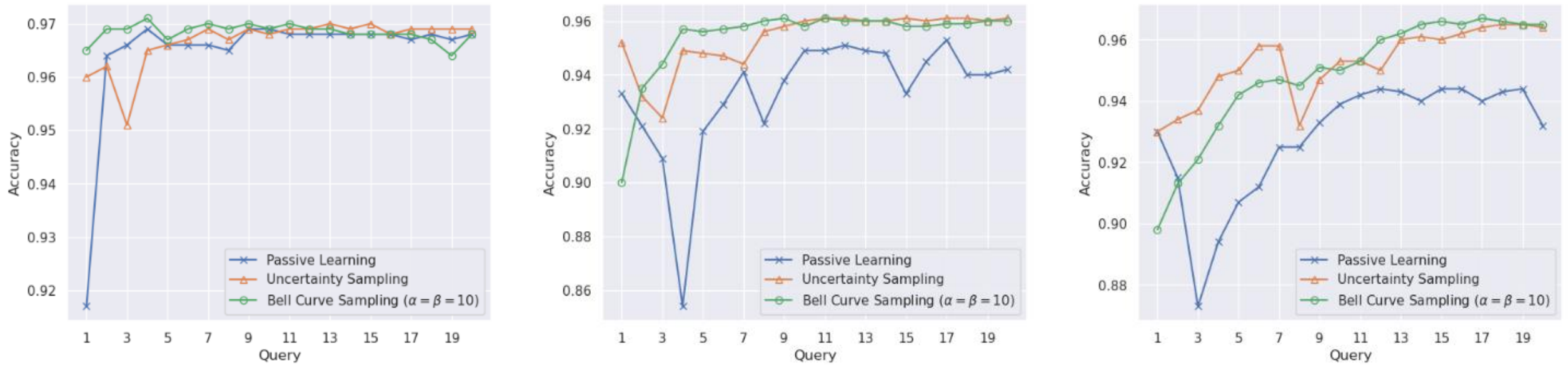

(a) Low AUR (b) Median AUR (c) High AUR

Fig. 6. Performance of various learning functions in blobs datasets of standard deviation (a) 1, (b) 3 and (c) 5.

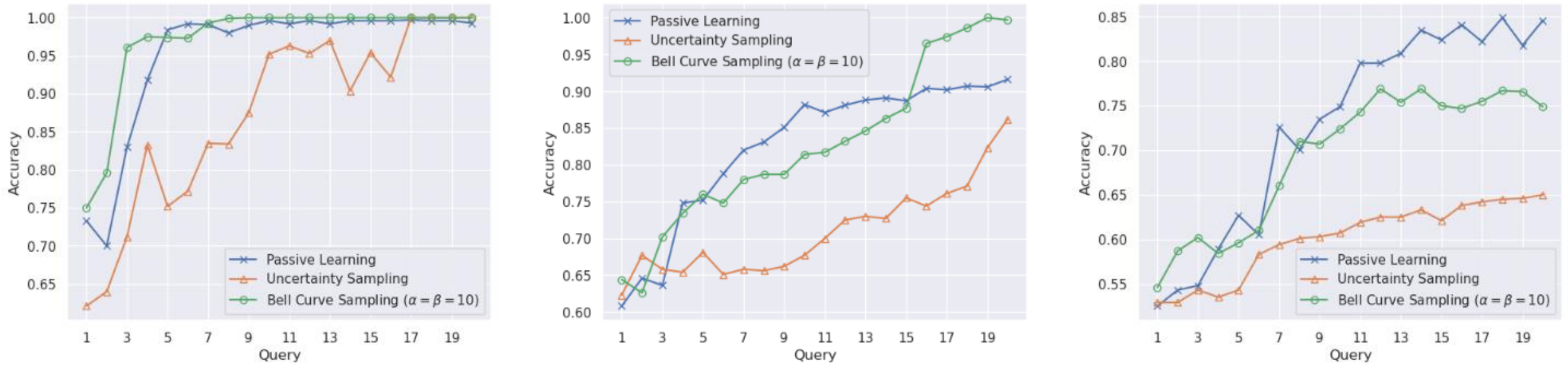

(a) Low AUR (b) Median AUR (c) High AUR

Fig. 7. Performance of various learning functions in circles datasets of circle factors (a) 0.5, (b) 0.8 and (c) 0.9.

Fig. 9 illustrates the impact of the parameters $\alpha$ and $\beta$ on the performance of bell curve sampling. As depicted in Fig. 3, increasing the values of $\alpha$ and $\beta$ will result in a steeper bell curve with a narrower width, implying that more weight will be given to instances closer to $\tilde{p} = 0.5$. In general, setting moderate to high values of $\alpha$ and $\beta$ can enhance the model's performance, while extremely high values will cause bell curve sampling to behave like uncertainty sampling.

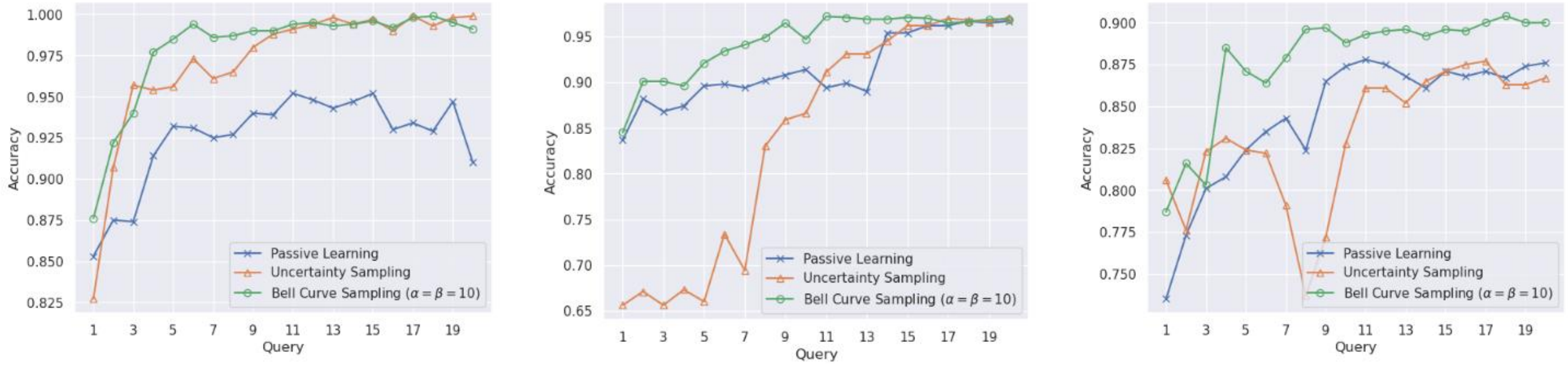

(a) Low AUR (b) Median AUR (c) High AUR

Fig. 8. Performance of various learning functions in moons datasets of Gaussian noise's standard deviation (a) 0.1, (b) 0.2 and (c) 0.3.

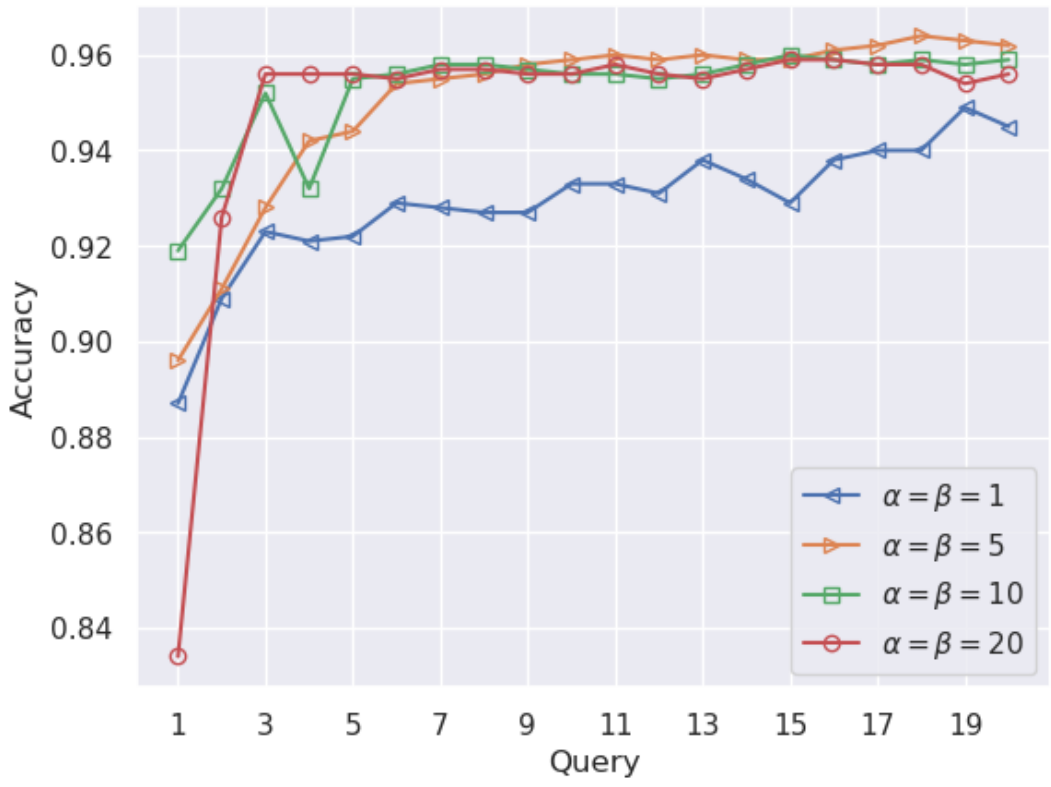

(a) Classification dataset

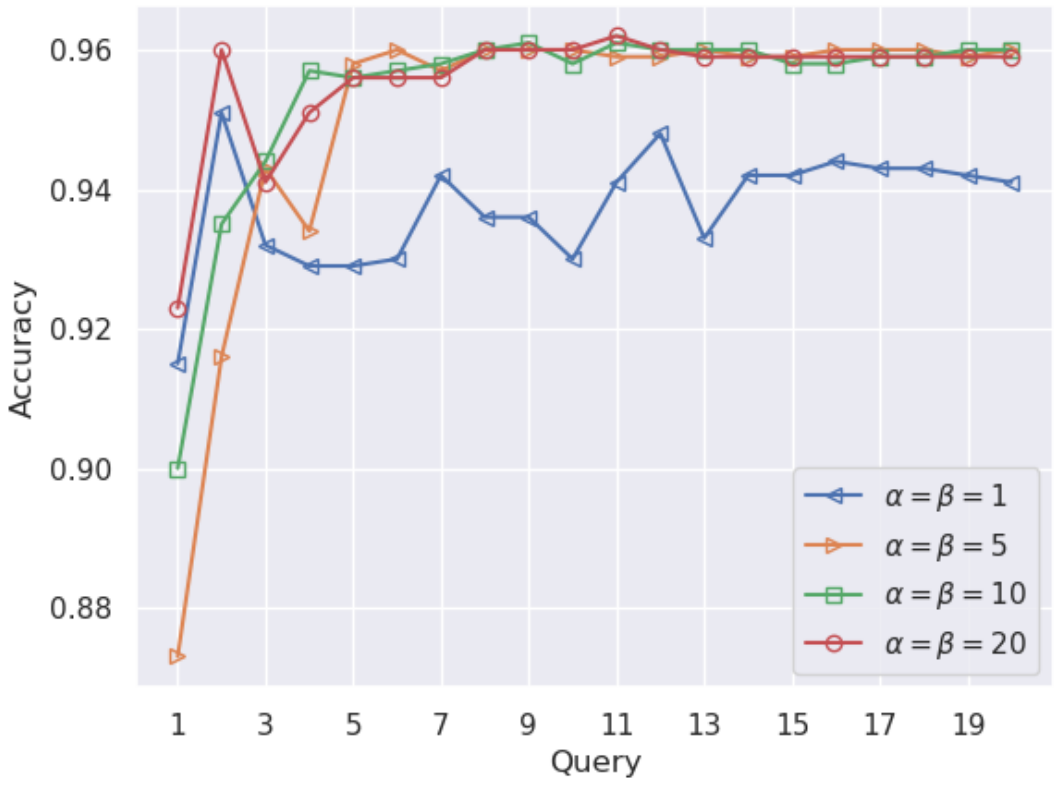

(b) Blobs dataset

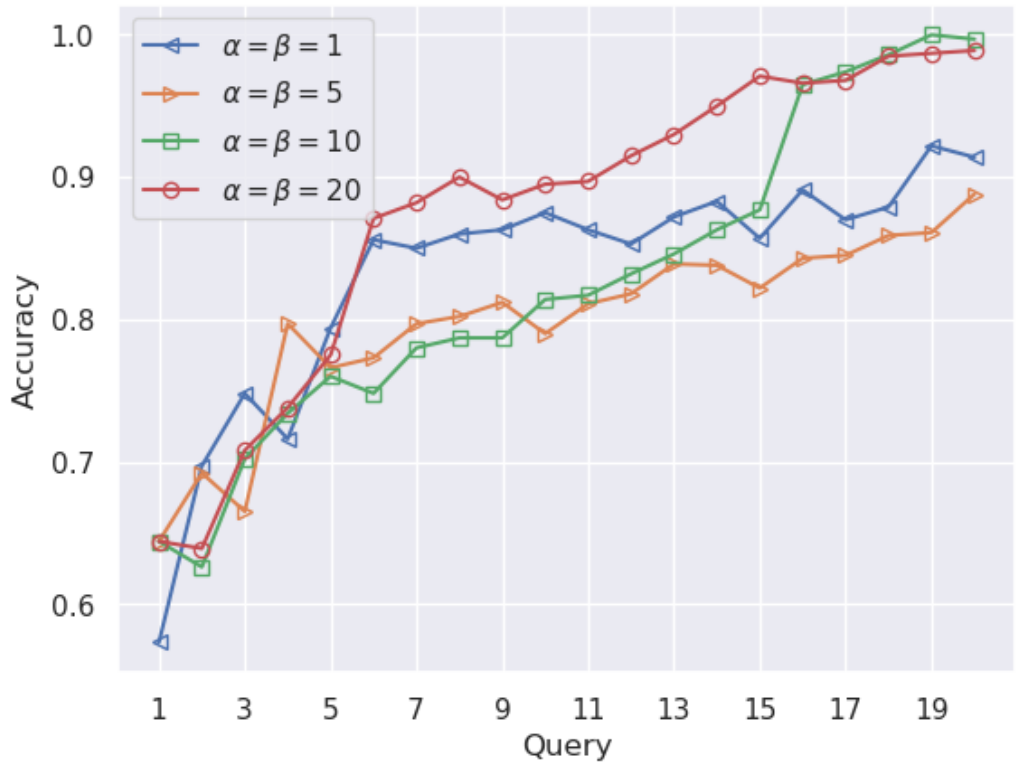

(c) Circles dataset

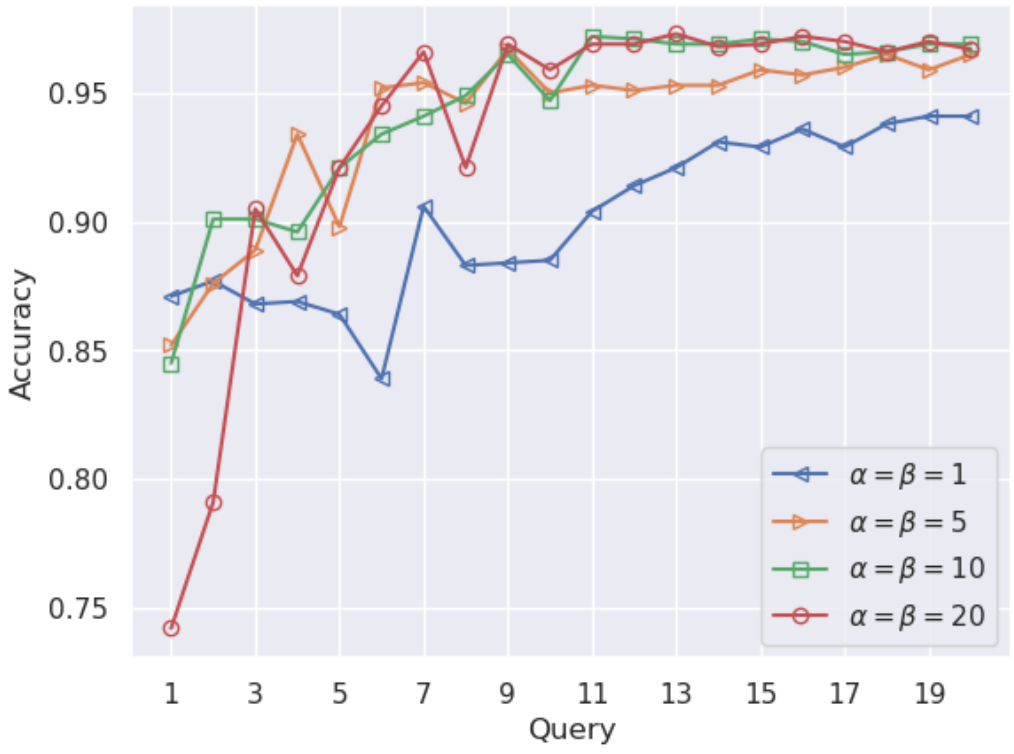

(d) Moons dataset

Fig. 9. The effect of the parameters $\alpha$ and $\beta$ to the performance of bell curve sampling.

## 5. Conclusion

Passive learning in typical machine learning is generally inefficient when the cost of acquiring labelled and unlabelled instances is different. To address this issue, active learning, especially uncertainty sampling, has been proposed. In principle, passive learning selects instances uniformly, while uncertainty sampling selects instances from the uncertainty region. Both sampling methods outperform each other in datasets of varying nature and AUR. Since selecting the best sampling method without prior knowledge about the datasets is difficult, we propose the use of bell curve sampling, which employs a bell curve weight function in selecting instances. Simulation results show that bell curve sampling performs better than passive learning and uncertainty learning for most datasets with diversified AUR values.

## Conflict of Interest

The authors declare no conflict of interest.

## Author Contributions

Zan-Kai Chong and Hiroyuki Ohsaki co-authored the research and the paper. Bok-Min Goi provided valuable input to enhance the paper. All authors had approved the final version.